\documentclass[10pt]{article}

\pdfoutput=1 
\usepackage[pagenumbers]{cvpr} 
\usepackage{times}
\usepackage{epsfig}
\usepackage{graphicx}
\usepackage{amsmath}
\usepackage{amssymb}
 \usepackage{booktabs}
 \usepackage{multirow}
 \usepackage{soul}
 \usepackage[table,xcdraw]{xcolor}

\begin{document}

\title{Unmanned Aerial Vehicle (UAV)-Based Mapping of Iris Pseudacorus L. Invasion in Laguna del Sauce (Uruguay) Coast.}

\author{Alejo Silvarrey\\
Departamento de Ingeniería\\
Universidad Católica del Uruguay\\
Av. Roosevelt Pda, 7 y 12, Maldonado, Uruguay\\
{\tt\small alejo.silvarrey@ucu.edu.uy}
\and
Pablo Negri\\
Instituto de Investigacion \\
en Ciencias de la Computacion (ICC),\\
UBA-CONICET.\\
Universidad de Buenos Aires, FCEyN\\
Computer Department, \\
Buenos Aires, Argentine\\
{\tt\small pnegri@dc.uba.ar}
}

\maketitle
\thispagestyle{empty}

\begin{abstract}
Biological invasions pose a significant threat to the sustainability of water sources. Efforts are increasingly being made to prevent invasions, eradicate established invaders, or control them. Remote sensing (RS) has long been recognized as a potential tool to aid in this effort, for example, by mapping the distribution of invasive species or identifying areas at risk of invasion. This paper provides a detailed explanation of a process for mapping the actual distribution of invasive species. This article presents a case studie on the detection of invasive Iris Pseudacorus L. using multispectral data captured by small Unmanned Aerial Vehicles (UAVs). The process involved spectral feature mapping followed by semi-supervised classification, which produced accurate maps of these invasive.

\end{abstract}

\section{Introduction}\label{sec:intro}
Natural resource agencies face a growing demand to prioritize the management of numerous non-native plants, specifically macrophytes, within freshwater ecosystems~\cite{kuehne2016}. Plants that are not native have the potential to change the physiochemical state of habitats, leading to adverse ecological and economic consequences~\cite{Xiong2023}. Also, Invasive alien species (IAS) and climate change, are identified as the top four drivers of global biodiversity loss~\cite{gentili2021invasive}. Thus, appropriate research strategies, funding mechanisms (for research, technology transfer and interventions such as management, control and prevention), and policies need to be developed and implemented~\cite{masters2010climate}.

Precise information regarding the spread of IAS is crucial for launching effective management initiatives~\cite{gentili2021invasive,masters2010climate}. Nevertheless, sampling endeavors are frequently inadequate and typically focus on only a limited portion of the actual distribution\cite{crall2015}. Thus, the significance of adopting innovative tools is evident for governments and environmental managers responsible for sustainable development. An instance of an emerging technology facilitating adaptable and cost-effective data collection for conservation and environmental management is Unmanned Aerial Vehicles (UAVs). While barrels are identified to applied UAV-based monitoring tools~\cite{walker2023barriers}, it is likely that some of them will lessen with time (e.g. technological and analytical barriers) as this technology continues to evolve.

The study of IAS in wetlands is of particular relevance, given the importance of this ecosystem. They provide essential resources for plants, animals, birds, and fish, which are exploited by hunter-gatherers as part of the seasonal economic cycle. Furthermore, they provide an optimal setting for the adoption and intensification of agriculture, offering greater stability and reduced risk during periods of environmental deterioration due to their provision of a reliable water supply~\cite{iriarte2001}. In some instances, they also represent a significant tourist attraction. With regard to coastal wetlands, both conservation and restoration may prove effective techniques for climate change mitigation~\cite{taillardat2020climate}.
Additionally, researchers have emphasized the necessity of effectively managing water body buffer zones to prevent the invasion of invasive alien species (IAS). As highlighted by Zelnik et al.~\cite{zelnik2015vulnerability}, the presence of native species enhances the benefits provided by buffer zones, including the mitigation of anthropogenic impacts on water bodies, stabilization of bank structures, and augmentation of biodiversity. Consequently, the control of IAS in buffer zones is crucial for the preservation of water body health.

Despite the chemical and mechanical control methods employed against IAS in different countries of the introduced range, this weed is still difficult to manage worldwide and continues to spread and cause significant negative environmental impacts~\cite{minuti2021,minuti2022,sandenbergh2021}. As governments and environmental managers strive to strike a balance between economic development and conservation in Uruguay and the region, understanding the ecological implications of IAS and developing effective management strategies have become imperative.

Obtaining aerial imagery becomes crucial when the target IAS exhibits a notable phenological characteristic, such as flowering. However, this needs the capability to acquire aerial images on short notice. Additionally, monitoring temporal changes in phenological development, such as achieving leaf-out earlier than the surrounding vegetation, demands multiple image-acquisition times throughout the growing season~\cite{hestir2008identification}. Capturing imagery at various points in the year or on specific dates at high spatial and temporal resolution poses a financial challenge for land managers when utilizing not-free satellite images. Thus, owing to the low cost and increasing capability of unmanned aerial vehicles (UAVs) for multispectral image acquisition, researchers have recently begun to explore this technology for mapping IAS~\cite{roca2022_1,da2023modeling_2,charles2021_3}.

This study aimed to assess the use of a small UAV (Mavic 3M) and a multispectral camera for remote sensing of IAS in wetlands at spatial and temporal resolutions to help achieve effective control measures. The present approach addresses all the challenges regarding this technique simply so it is feasible to be applied in different wetlands. UAV-based remote sensing is a powerful tool capable of complementing the current pool of monitoring and mapping alternatives.

\subsection*{Target: Iris Pseudacorus L.}

The Iris pseudacorus L. allso know as a Yellow Flag Iris (YFI) is a perennial flowering plant known for its striking yellow flowers and distinctive sword-shaped leaves. It belongs to the iris family, Iridaceae, and is native to Europe, Western Asia, and North Africa~\cite{engin1998,sutherland1990}. Due to its hardiness and ability to thrive in wet conditions, the plant has been introduced to other regions for its ornamental value, including South America, where it is considered invasive~\cite{masciadri2010,gervazoni2020,EEI_UY}.

The YFI is characterized by its tall, erect stems that can reach heights up to 1 to 1.5 meters (3 to 5 feet). The leaves are long and narrow, resembling swords, a common feature of iris plants. The most notable feature of the YFI is its yellow flowers, which typically bloom in late spring to early summer~\cite{sutherland1990}. The flowers have a distinct appearance, with three upright petals and three drooping sepals.

As was mentioned before, While valued for its aesthetic appeal, the YFI has raised concerns in some areas as it can become invasive and outcompete native vegetation. Its ability to form dense stands along watercourses and wetland areas may impact local ecosystems, which include changes in habitat structure, alteration of nutrient cycling, and impacts on other plant and animal species.

Previous works on mapping I.pseudacorus with UAVs have explored two methods of analysing the imagery acquired: (1) visual analysis of the individual images and the orthomosaic image created from them and (2) supervised, pixel-based image classification of the orthomosaic image~\cite{hill2017}. The research conclude that manual interpretation of the UAV-acquired imagery produced the most accurate maps of yellow flag iris infestation. This outcome represents an exciting advance in YFI mapping. However, a more effective and efficient approach may facilitate the task.

\subsection*{Study Area: Laguna del Sauce (Maldonado/Uruguay)}
Laguna Del Sauce is located 20 km north-west of Punta Del Este, the main tourist city in Uruguay and 100 km east of Montevideo, the capital city of Uruguay. It is covers an area of 40.5 km2, is 12 km long and 6 km wide, and its depth varies between 3 and 5 meters~\cite{barruffa2020}.

\begin{figure*}[t]
 \centering
 \includegraphics[width=0.9\textwidth]{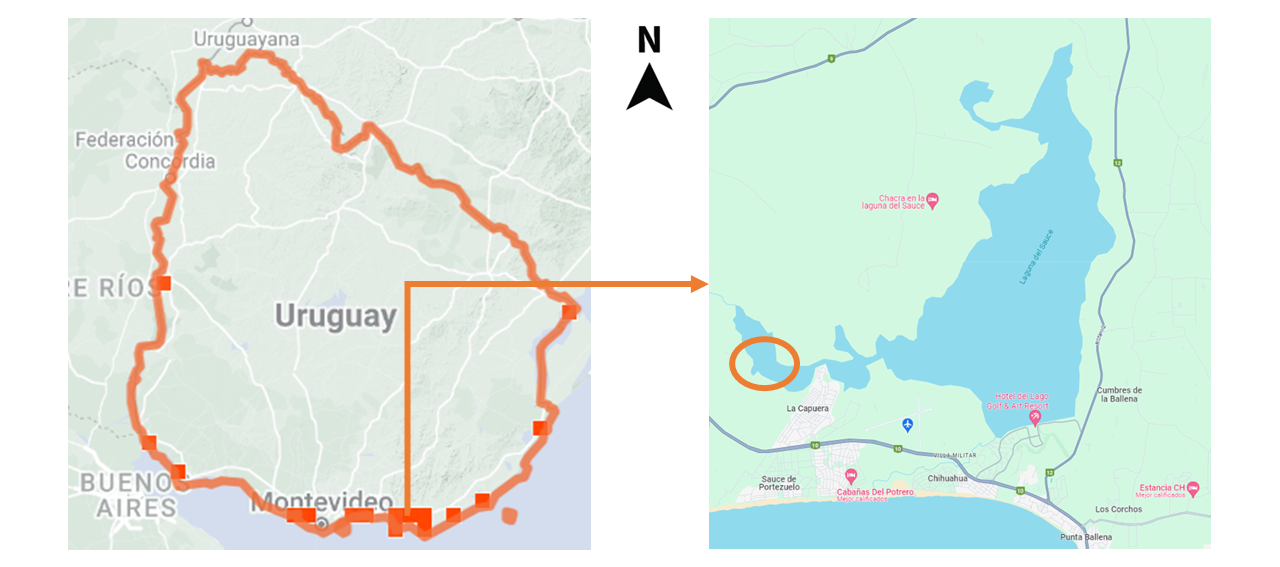}
 \caption{On the left of the figure, the distribution map of the Yellow Flag Iris in Uruguay is shown. The study area in Laguna del Sauce (Maldonado/Uruguay) is indicated on the right. Source: Developed by the authors based on the webpage NaturalistaUY\cite{NaturalistaUY} and the reserach conducted by G. Minuti~\cite{minuti2022}}
 \label{distribution map}
\end{figure*}

In 2021, I. pseudocorus was detected in the wetlands of the Laguna del Sauce catchment, specifically in the coastal area on the southern side of Laguna del Potrero (subsystem of Laguna del Sauce). The patches detected cover an area of 0.3 hectares at its highest density, with specimens distributed in very isolated patches along approximately ten linear kilometers of shoreline (Lagunas del Potrero and de los Cisnes). In the focus of the highest density, the specimens are established both in the floodplain and on floating islands formed by water hyacinths (Eichornia crassipes).

The research was conducted in EL Chajá, a plant nursury located on the Laguna del Potrero where the YFI patches occurred (\ref{distribution map}). On this area others reserch related to YFI management are conducted by the owner and national and local authorities. 

\subsection*{Management measures of YFI in Laguna del Sauce Catchment}
The introduction and spread of YFI is a serious threat to ecosystem structures, functions and services. At enormous cost and effort, attempts are being made to stop and reduce the ongoing expansion in uruguayan wetlands to protect the regional and native biodiversity. Still, management of invasive alien species in particular is often not successful~\cite{schmiedel2016}.

To mapping YFI on Laguna del Sauce, the National Alien Species Control Committee encourage the use of iNaturalist, a comprehensive global platform for documenting species observations data~\cite{Expediente}. This popular community science platform enables people to contribute valuable data and records undergo a rigorous community-based identification process. Thus, an observation is regarded as research grade when it has location, date, media evidence, and two or more suggested identifications, more than two-thirds of which agree at a species level.

This platform was used by other researchers for biodiversity monitoring on a global scale~\cite{nugent2018inaturalist}. Notably, in the case of Uruguay, the coastal zone where YFI occurs has been extensively sampled by citizen efforts~\cite{grattarola2023}, offering valuable data for studying the invasion dynamics of this species.

Although the species has pockets in the country where eradication is challenging, it is understood that the situation in Laguna del Sauce presents conditions that make it approachable for control, in which it is possible to consider successful results in the medium term. The size of the focus justifies this, the functional relevance of the affected wetland, and the institutional participation in solving the problem.
The Control Plan was developed within the framework of the YFI Control Working Group, which operates within the scope of the Catchment Commission~\cite{Expediente}. It is structured around the three axes established in the IAS Response Protocol~\cite{CEEI2018}: 1- Research, 2- Base Line and Control and 3- Divulgation.

Based on the YFI Control Working Group approach, monitoring the spread and studying the mechanisms and strategies of invasions require spatially-explicit approaches~\cite{Expediente}. Remote sensing technologies provide the necessary information to develop cost- and time-effective solutions for assessing current and future invasion processes over large geographical extents. Also, it is a powerful tool capable of complementing the current pool of monitoring alternatives as iNaturalist.

\section{Methodology}

A lake in Maldonado (Uruguay) were used to evaluate the performance of the
methods used to quantify the location and extent of YFI.

\subsection*{Image acquisition and processing}
In order to develop a map of the YFI, a proximity multispectral approach has been adopted. This was achieved by means of a DJI Mavic 3M Professional aircraft (DJI Sciences and Technologies Ltd) equipped with a multispectral camera. 
The DJI Mavic 3M is a fully autonomous quadcopter that is capable of capturing high-resolution aerial images. 
The multispectral camera has four multispectral sensors: green, red, red edge and NIR band with a resolution of 4x5 MP. 
Furthermore, the device incorporates an RGB camera with a 20-megapixel resolution and a sunshine sensor. 

To survey the site, an automatic flight mission was planned setting a 120 m altitude. This value was adopted to cover as much as possible per image, reduce the flight time and consider the flight restriction by local regulation. The flight cover 62.000 m2 and the multispectral camera was set up to capture 111 images. The resulting pixel resolution is 5 cm by 5 cm. It is relevant to mention that the flight took 4´12". According to Laguna Del Sauce meteorological station, during the survey, atmospheric conditions are clear skies and low wind velocity (less than 25 km/h). Data collected under cloudy and windy conditions may introduce variability in solar irradiance so the survey was conducted under these conditions. Also, flights should be conducted at times to avoid sunglint. The presence of sunglint is best evaluated locally at the time of the intended flight, thus, the survey took place Mid-morning based on previous experiences on the study region\cite{barruffa2021monitoring}.

The mosaic was created using one of the most common software programs, Pix4D Mapper Pro. This software is capable of applying radiometric corrections based on image metadata. Additionally, the software aligned the camera bands in order to obtain a correlation in each pixel. The processing stages in Pix4D Mapper include initial processing, point cloud and mesh, and orthomosaic. For each of these stages, all options were set to the default.

An area of YFI was delineated using 30 x 30 cm cardboard panels. These panels are visible on images, and geo-localized polygons can be created using them as a reference on QGIS software. Thus, the portion of each band orthomosaic lacking YFI was removed using the function "extract by mask," where the authors first created the features using the panels. The panel approach was employed in other research with Unmanned Aerial Vehicles\cite{murfitt2017}.

\subsection*{Yellow Flag Iris Detection Pipeline}

The objective of the YFI detection procedure is to generate a binary mask, denoted by the symbol $\mathbf{Y}$, for the input image. The mask assigns a value of one to the pixels within a sector containing the specified plants, and zero to all other pixels. To characterize the yellow flag iris in the image, sectors with the presence of the plant are identified manually through a field survey. A couple of markers are planted on the ground for this purpose, which can be visualized later in the UAV image. 
The result is a binary mask, \textbf{K}, which will be used as the sole \textit{a priori} knowledge to detect the yellow flag iris in other image sectors. The detection strategy follows the following steps:

\begin{itemize}
    \item First hypothesis mask $\mathbf{H}$: this step generates a binary mask $\mathbf{H}$ by analyzing the range of multispectral channel values of the input image within the labeled mask $\mathbf{K}$. Mask $\mathbf{H}$ defines a group of $n_H$ sets of connected components $R_i$, with $i=1,...,n_H$.
    \item Filtering $R_i$ by flowers presence: this stage estimates the presence of yellow flowers inside each set $R_i$ using both, \textit{green} and \textit{red} channels. A yellow flower is defined by a local maxima (peak) $m_j$ at the plane generated by this channel combination. All the local maxima positions inside the set $R_i$ are grouped at $\mathbf{m}^i=\{ m_j \}_{j=1,...,q_i}$. Those sets $R_i$ with fewer flowers than a specific threshold are eliminated from the mask $\mathbf{H}$.
    \item Finally, the output mask $\mathbf{Y}$ is generated by remaining sets $R_i$.
\end{itemize}

The following section details each pipeline step and presents the results of detecting YFI regions.

\section{YFI Regions Detection}
\subsection*{\textit{A priori} knowledge}

The procedure for detecting the YFI is semi-supervised. Prior knowledge about the location of the flowers is required to capture information about the multispectral image range of values, texture, etc. The target regions containing yellow flowers are defined with a binary mask \textbf{K}.

In our case, $\mathbf{K}$ contains $L=15$ labeled sectors with flowers.

\begin{figure}[h]
\centering
  \includegraphics[width=0.9\columnwidth]{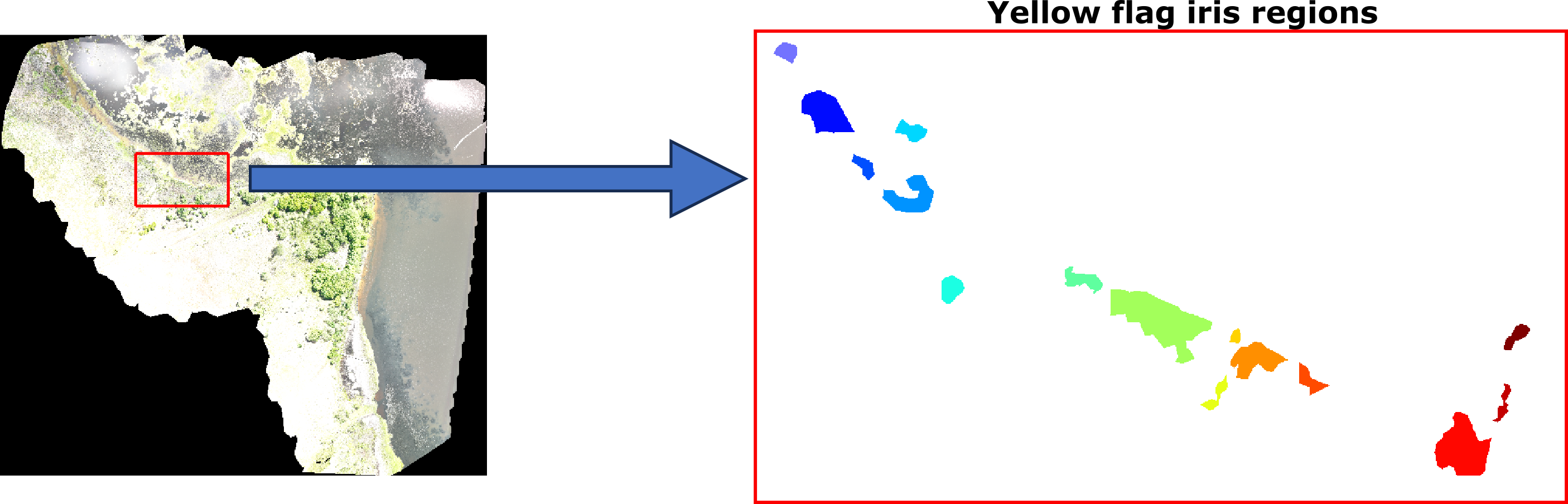}
  \caption{This figure shows the original image captured on the left, and the red rectangle defines the manual labeled YFI regions sector. The right image shows in detail the sets of connected-pixels in the mask $\mathbf{K}$, with a different color for each disjoint set of the list $[t^{(i)}]_{i=1,...,L}$.}
  \label{fgr:maskK}
\end{figure}

Each region is an independent subset of connected pixels, and defined as $[t^{(i)}]_{i=1,...,L1}$
The subsets have an average 6.700 pixels.
The masked area in $\mathbf{K}$ represents the $0.087$ \% of the overall image. Fig. \ref{fgr:maskK} shows the sector in the original image where the regions of YFI were manually labeled.

The following hypotheses are assumed to characterize the presence of yellow flowers using multispectral images:

\begin{itemize}
    \item [H1] The regions around yellow flowers have similar range values of multispectral channels. 
    \item [H2] Yellow flowers are local peaks in GREEN and RED multispectral channels.
    \item [H3] Yellow flowers grow in proximity to one another, rather than in isolation.
\end{itemize}

\subsection*{Multispectral channels}\label{sec:channels}

The detection of YFI is accomplished through the use of multispectral images captured from a UAV. This section provides a detailed account of the data provided by the instrument and the conversions applied to enhance the detection of chlorophyll.
The utilization of multispectral channels in this study is justified by their enhanced robustness in the presence of varying lighting conditions and shadows, among other factors.

The RED and GREEN channels of the multispectral image enhance the images of the flowers, as shown in Figure \ref{fgr:umbrales}. 

\begin{itemize}
    \item GREEN: the spectral reflectance measurements acquired in the green (visible)
    \item RED: the spectral reflectance measurements acquired in the red (visible)
\end{itemize}

Sectors containing plants can be identified by a combination of channels that measure chlorophyll content.
The Normalized Difference Vegetation Index (NDVI) determine changes in chlorophyll content and crop growth status. NDVI is widely used in various applications, including precision agriculture and vegetation stress detection\cite{zhang2021novel}.

    \begin{equation}
        NVDI = \frac{(NIR-RED)}{(NIR+RED)}
    \end{equation}
    where NIR is near-infrared regions.

Two additional channels has the ability to enhance chlorophyll indexes (CI): green chlorophyll index (CI-GREEN), and red edge chlorophyll index (CI-EDGE) \cite{gitelson2003relationships}. 
They are based on a near-infrared band and a band located in the medium chlorophyll absorption domain.
    \begin{equation}
        CI-GREEN = \frac{NIR}{GREEN} - 1.
    \end{equation}
    \begin{equation}
        CI-EDGE = \frac{NIR}{RED EDGE} - 1.
    \end{equation}

\subsection*{Region Hypotheses Generation using Mathematical Morphology}

The purpose of this subsection in the detection pipeline is to eliminate non-plant sectors in the image, such as water, sand, and dirt. 

The aim is to retain regions in the image with vegetation as hypotheses.
Thus, mathematical morphology is used on the multispectral channels described in section \ref{sec:channels}. These channels enhance the presence of pixels with chlorophyll. Although it may seem like a straightforward task, it is essential to be cautious not to remove areas with yellow flowers. These areas would not be included in the subsequent process and would no longer be detected if this is done.

Mathematical morphology is a widely used tool in digital processing to handle digital images \cite{serra:84}. It can obtain high level segmentation features using simple operations while requiring low memory.
Considering that multispectral images are huge, mathematical morphology is the best solution.

\begin{figure}
\centering
  \includegraphics[width=0.9\columnwidth]{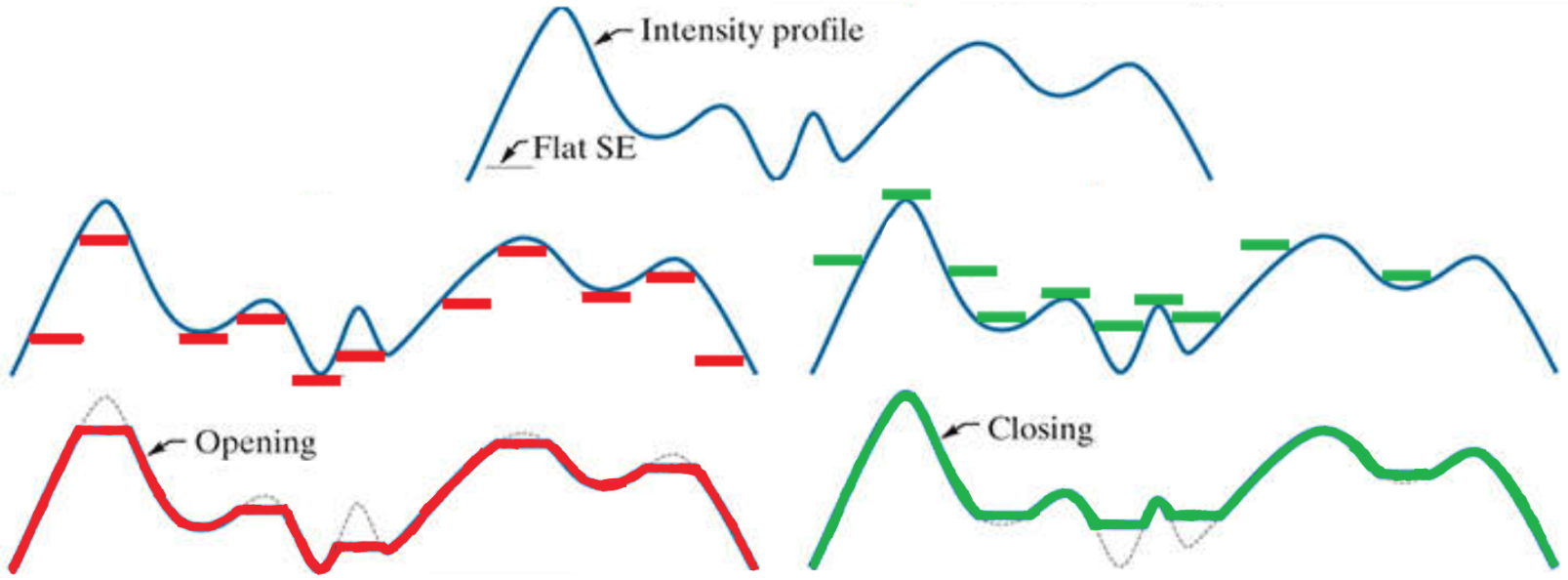}
  \caption{The figure depicts how OPEN and CLOSE morphological operations perform on 1D signal (not binary). Source: image modified from \cite{gonzalez:2018}.}
  \label{fgr:morfologia}
\end{figure}

We use two basic operations of the mathematical morphology: OPEN and CLOSE.
Fig. \ref{fgr:morfologia} depicts the result of a structuring element on the profile of a 1D signal values. While the OPENING operation preserves minimal value and is pushed up underneath the signal, the CLOSING operation keeps maximal values and is pushed down along the top of the signal.

First, to characterize the pattern of YFI regions, we choose one of the subsets, the $t^{(9)}$, which have a high number of yellow flags.

\begin{figure}
\centering
  \includegraphics[width=0.9\columnwidth]{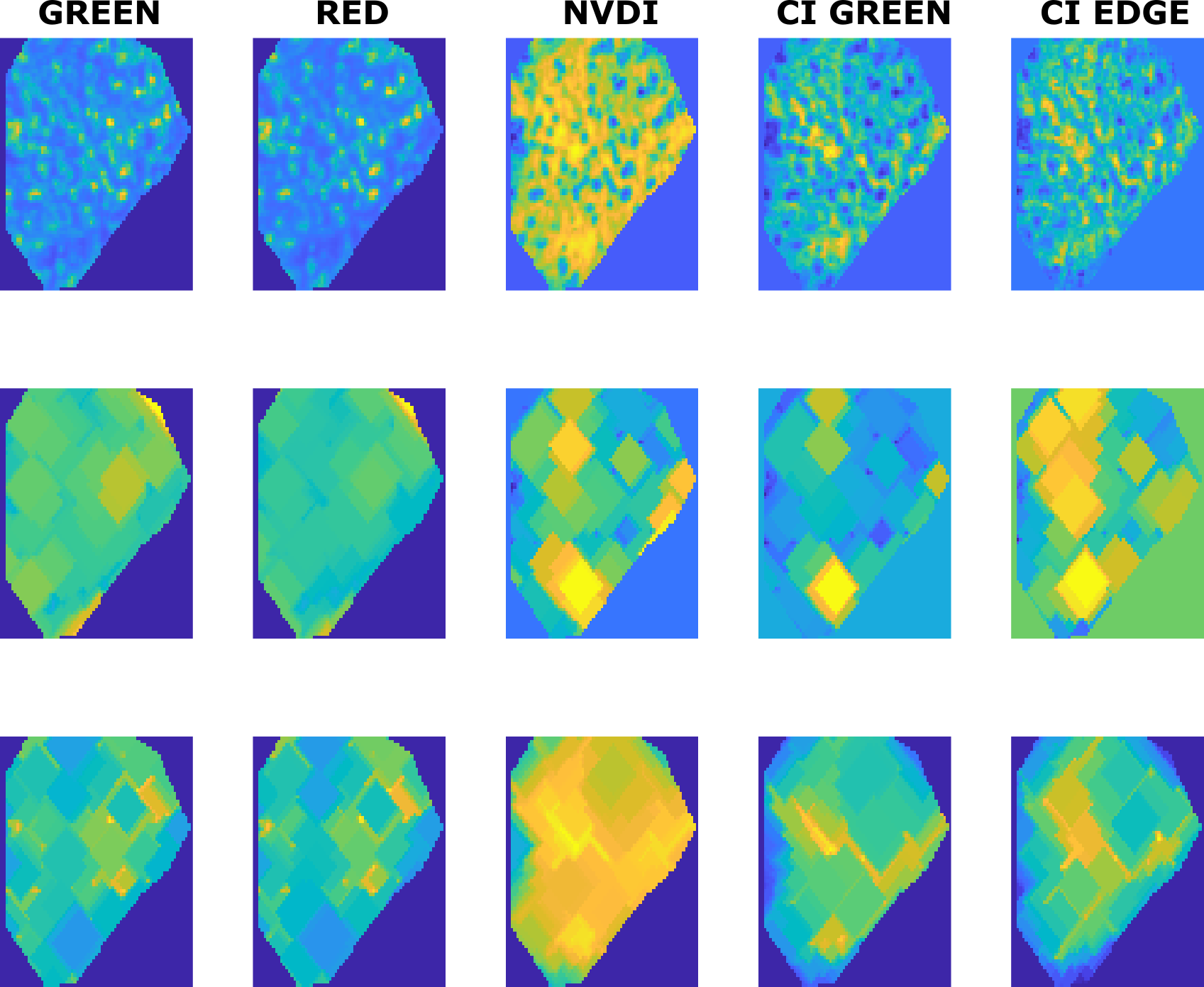}
  \caption{The figure shows the five channels employed to detect the first hypotheses with YFI plants. The second and third rows depict the result of the OPEN and CLOSE operations, respectively, using the same kernel.}
  \label{fgr:umbrales}
\end{figure}

Fig. \ref{fgr:umbrales} presents the OPEN and CLOSE operation over the five channels on the $t^{(9)}$ subset.
Both operations employ a flat diamond shape of $15 \times 15$ pixels size structuring element.

The subsequent step delineates a range of values within subset $t^{(9)}$, which encompasses the OPEN and CLOSE operations. This range of values serves to threshold the OPEN and CLOSE maps, thereby binarizing the values of each channel. To circumvent the influence of outliers, the thresholds are calibrated to retain 95\% of the activated pixels (with one value in the binary mask) within a subset $t^{(9)}$. In a formal sense, the thresholds for the OPEN operation are defined as the minimal value, denoted as $\underline{T}^{OPEN}$, and the maximal value, denoted as $\overline{T}^{OPEN}$.
Similarly, the CLOSE operation also delineates analogous thresholds, as $\underline{T}^{CLOSE}$, and $\overline{T}^{CLOSE}$. These thresholds are applied to each channel.

Applying $(\underline{T}^{OPEN},\overline{T}^{OPEN})$ and $(\underline{T}^{CLOSE},\overline{T}^{CLOSE})$ thresholds on GREEN channel, it generates the following binary mask:
\begin{eqnarray}
    \mathbf{B}^{OPEN}_{GREEN} & = & (GREEN \geq \underline{T}^{OPEN}_{GREEN}) \cap (GREEN \leq \overline{T}^{OPEN}_{GREEN}) \\
    \mathbf{B}^{CLOSE}_{GREEN} & = & (GREEN \geq \underline{T}^{CLOSE}_{GREEN}) \cap (GREEN \leq \overline{T}^{CLOSE}_{GREEN}) 
\end{eqnarray}

\begin{figure*}[t]
\centering
  \includegraphics[width=0.9\textwidth]{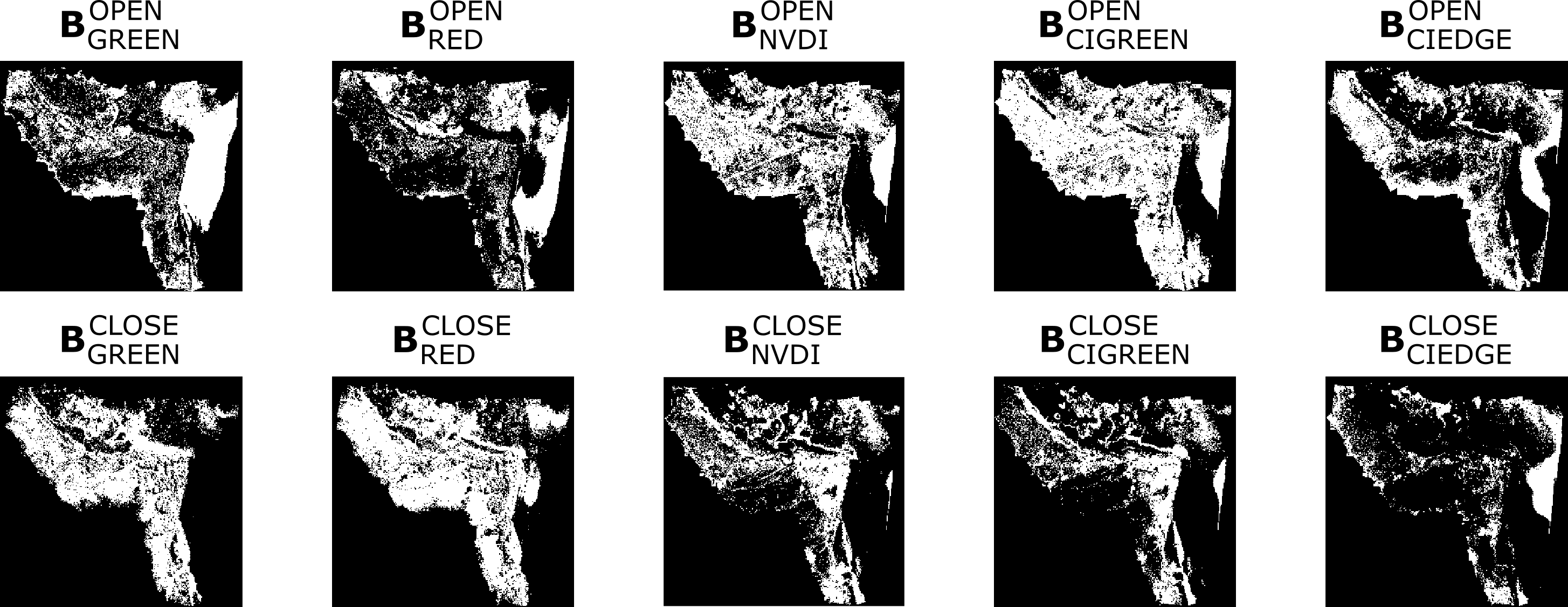}
  \caption{The figure shows binarization on the five channels using the thresholds found by the OPEN and CLOSE channels maps on the overall image.}
  \label{fgr:mascarasCanales}
\end{figure*}

The remaining channels are operated in a manner analogous to that described above for the computation of the other binary masks. Fig.\ref{fgr:mascarasCanales} illustrates the binary masks for all channels, with the thresholds for the OPEN and CLOSE operations superimposed. It can be observed that the OPEN masks and CLOSE masks exhibit complementary behavior. In general, CLOSE masks filter water pixels, while OPEN masks retain terrain and vegetation pixels.
Some masks, such as ${B}^{OPEN}_{NVDI}$ and ${B}^{OPEN}_{CIGREEN}$, are not discriminant and validate most of the pixels, while others exclude a very high number of pixels.
Mask ${B}^{CLOSE}_{CIEDGE}$ is an example of such a filter, which preserve regions of interest within the range of chlorophyll values of the yellow flag, others filters. 

To obtain the final masks, $\mathbf{B}^{OPEN}$ as the intersection of all the OPEN masks, and $\mathbf{B}^{CLOSE}$ as the intersection of all the CLOSE masks, are combined in the first hypothesis mask $\mathbf{H}$: 
\begin{eqnarray*}
    \mathbf{B}^{OPEN}&=& {B}^{OPEN}_{GREEN} \cap {B}^{OPEN}_{RED} \cap {B}^{OPEN}_{NVDI} \cap {B}^{OPEN}_{CIGREEN} \cap {B}^{OPEN}_{CIEDGE} \\
    \mathbf{B}^{CLOSE}&=& {B}^{CLOSE}_{GREEN} \cap {B}^{CLOSE}_{RED} \cap {B}^{CLOSE}_{NVDI} \cap {B}^{CLOSE}_{CIGREEN} \cap {B}^{CLOSE}_{CIEDGE} \\
    \mathbf{H}&=&{B}^{OPEN} \cap {B}^{CLOSE}
\end{eqnarray*}

\begin{figure}
\centering
  \includegraphics[width=0.85\columnwidth]{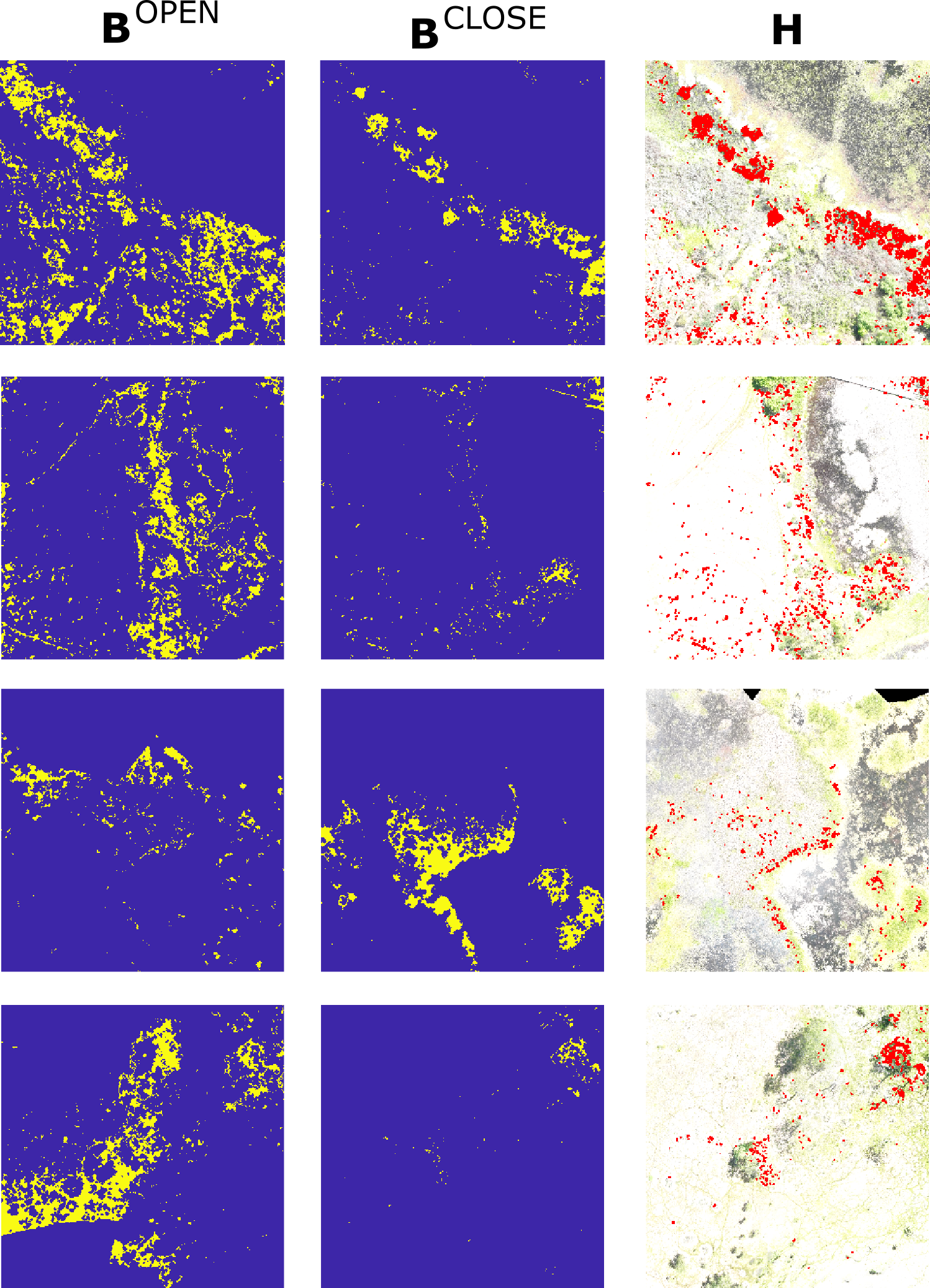}
  \caption{The final result $\mathbf{H}$ is the intersection of OPEN and CLOSE masks.}
  \label{fgr:mascaraFinal}
\end{figure}

Fig. \ref{fgr:mascaraFinal} shows the final yellow flags hypothesis regions.
The number of hypothesis regions in \textbf{H} is 10,517.
The average area is 42 pixels, the maximum area is 32,712 pixels and the minimum area is 1 pixel.
After a rapid filtering of small and thin regions with the OPEN operation on the binary mask using a square $5 \times 5$ kernel size, the number of regions was reduced to 2.344.

To connect neighbor regions and fill holes in the binary mask, DILATE and CLOSE operations with disk kernels of $18 \times 18$ kernels size are then applied. The final number of regions in \textbf{H} reaches 524.

The next section estimates the number of yellow flowers inside each region set to filter false hypotheses.

\subsection*{Detecting Flowers}

\begin{figure}
\centering
  \includegraphics[width=0.65\columnwidth]{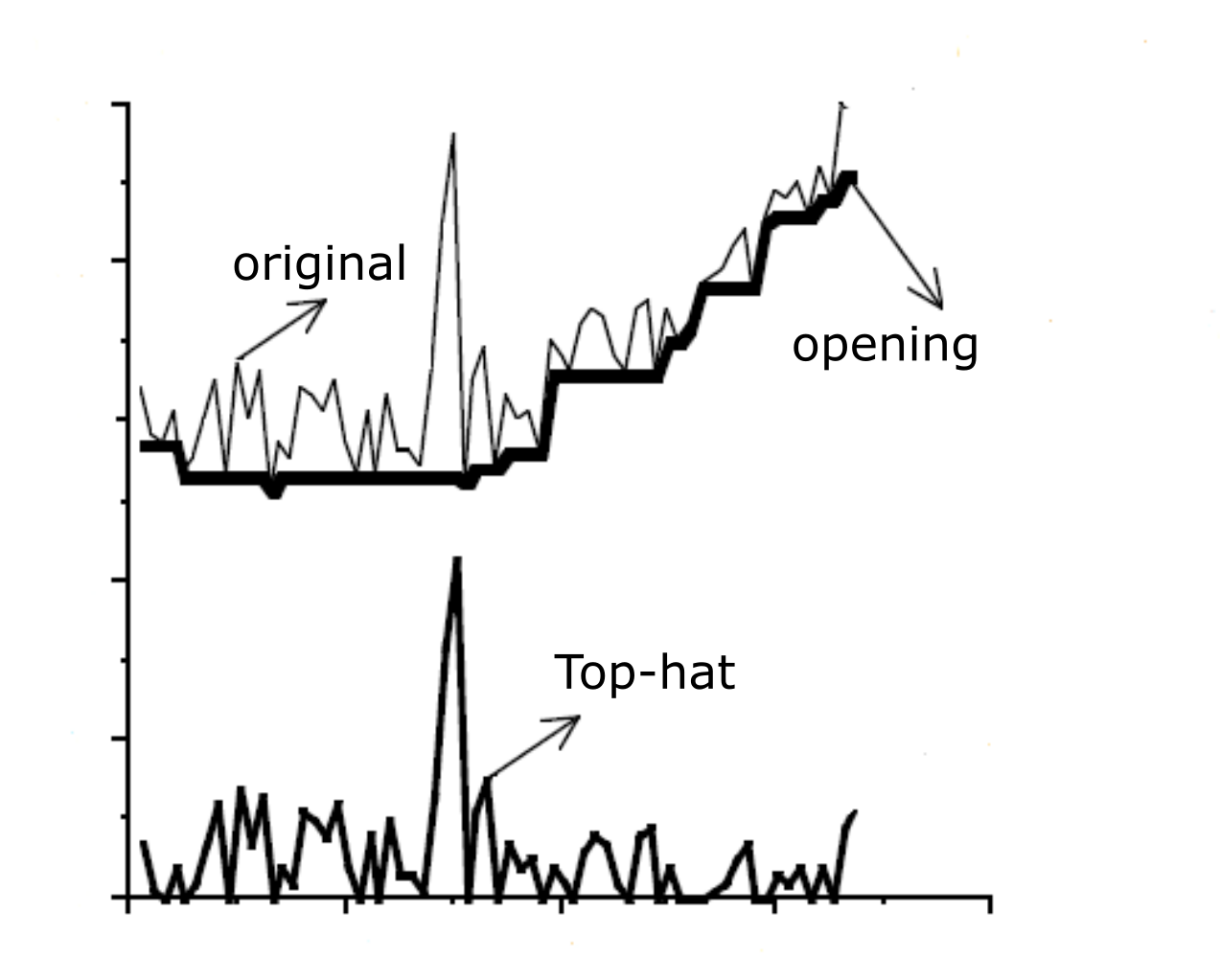}
  \caption{Top-Hat operation (Picture modified from \cite{SERRATOPHAT}).}
  \label{fgr:tophat}
\end{figure}
The second hypothesis to identify yellow flowers involves detection of peaks in GREEN and RED channels using Top-Hat filtering~\cite{soille:99}.
Top-Hat filtering of image $I$ is the difference between $I$ and its opening. 
Fig. \ref{fgr:tophat} shows the effect of the filtering in a 1D signal, where it can be seen that signal peaks are extracted independently from their intensity level.
This example can be thought of as the profiles of GREEN and RED channels, where yellow flowers correspond to the local maxima values (peaks).
Top-hat filtering preserves these maxima and normalizes the lower bound of the signal.

\begin{figure}
\centering
  \includegraphics[width=0.65\columnwidth]{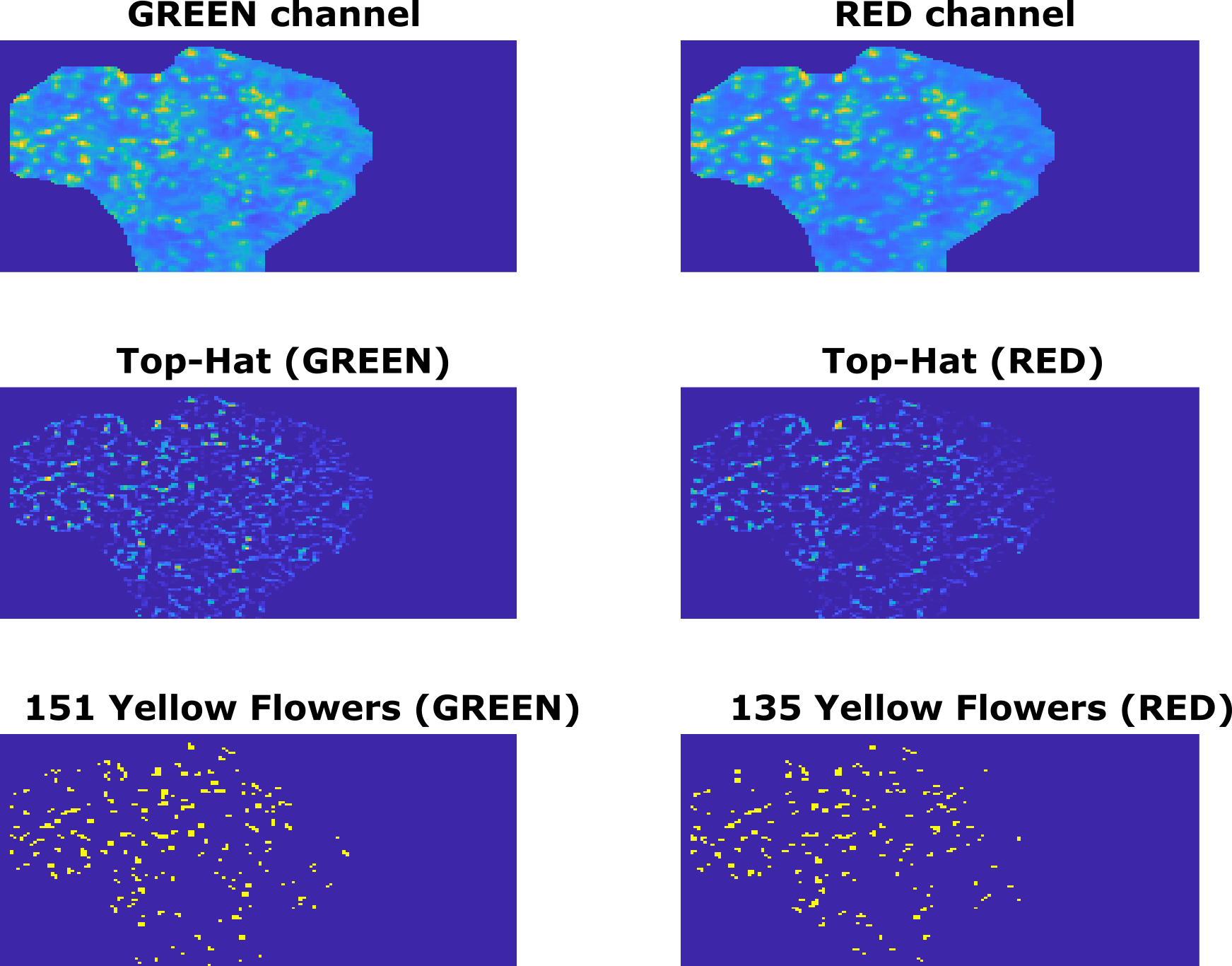}
  \caption{Top-Hat operation and filtering in GREEN and RED channels. The last row shows the binary peaks which correspond to yellow flowers.}
  \label{fgr:tophatreg}
\end{figure}

The methodology for identifying yellow flowers in each labeled region is straightforward.
First, apply a Top-Hat filter to the GREEN channel using a structuring element with a cross shape and a size of 3x3 pixels.
This structuring element will preserve the peaks of the yellow flowers.
The next step involves setting a threshold of $T^{(i)}_{green}$ to preserve only the highest local maxima, which represents 1
The average of the thresholds calculated at each labeled region is used to compute $T^{a}_{green}$ in the following step.
Then, the Top-Hat filtering of the GREEN channel is thresholded with $T^{a}_{green}$ in the final step, and the number of detected local maxima at each labeled region is counted as yellow flowers $Q^{(i)}_{green}$.
The same methodology is applied to the RED channel to obtain the number of detected yellow flowers inside each region $Q^{(i)}_{red}$.

The quantity of yellow flowers within a region will be used later to validate the region hypothesis in mask $H$.

\subsection*{Hypotheses regions validation}

This section validates hypotheses regions $h_j$ in mask $H$ using: thresholds of local maxima $(T^{a}_{green},T^{a}_{red})$, and the amount of yellow lowers per region $(Q^{(i)}_{green},Q^{(i)}_{red})$

The procedure starts applying thresholds $(T^{a}_{green},T^{a}_{red})$ at both GREEN and RED Top-Hat filtered channels inside each hypothesis region $h_j$.
Values $q_{green}^j,q_{red}^j$ are the amount of peaks inside region $h_j$.
Hypothesis region $h_j$ becomes yellow flower detection region $f_k$ if and only if both quantities are more significant than the minimum of lists $(Q^{(i)}_{green},Q^{(i)}_{red})$.

Regions $f_k$ with an important number of peaks, populate the final mask $F$ which is the output of the yellow flowers detection procedure.
 
\subsection*{Detection results}

\begin{figure}
\centering
  \includegraphics[width=0.95\columnwidth]{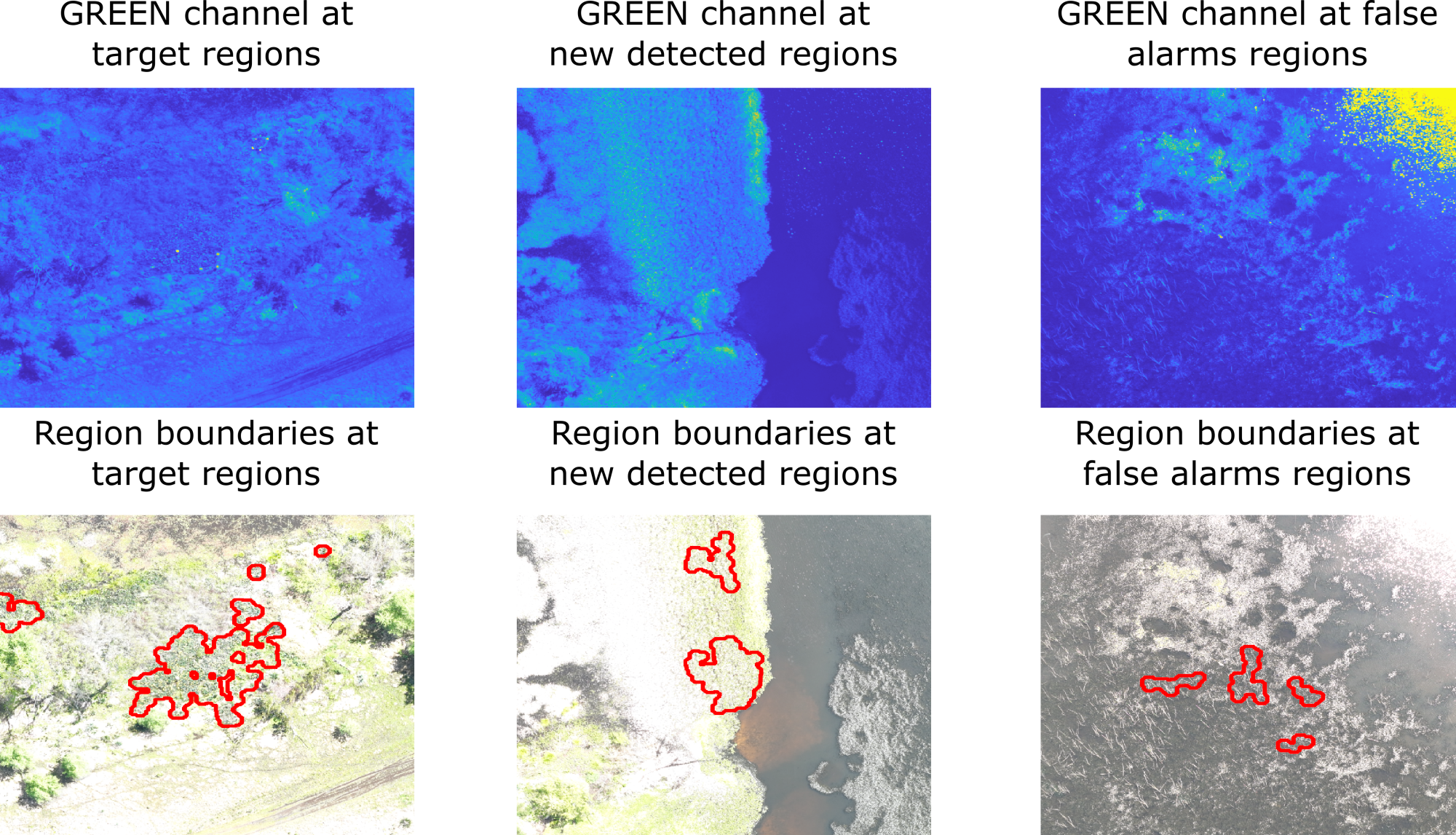}
  \caption{Three examples of detected yellow flowers in an independent image.}
  \label{fgr:results}
\end{figure}

The procedure for detecting regions with yellow flowers was applied to a new set of multispectral images captured on the same day, but 90 minutes after the images used to define the thresholds. This was done to validate the robustness of the semi-supervised procedure.

Figure \ref{fgr:results} shows three examples of the detected regions.
The left column depicts the boundaries of the detected regions in the sector adjacent to the area where yellow flowers were previously detected.
The first row shows the texture and values of the GREEN channel. 
The central column illustrates the borders of recently identified regions in the lake, and the right column shows the three false alarm regions.
False alarms were positioned in the middle of the water, making them easy to filter by an operator.

\section{Discussion}
IAS can harm ecosystems, leading to biodiversity loss and changes in ecosystem functions. Effectively addressing these impacts necessitates understanding these species' spatial distribution and extent of infestation. Although in situ inventories or aerial photo interpretation can be employed for data collection, these methods are labor-intensive, time-consuming, and often incomplete, particularly in large or hard-to-reach areas\cite{nininahazwe2023mapping}. Remote sensing emerges as a promising approach for mapping IAS, offering a more efficient means for developing comprehensive management strategies. The main objective of this study was to develop a mapping method for a major IAS observed in Wetlands of Maldonado (Uruguay), namely Yelow Flag Iris (Iris Pseudacorus L.).

This research shows that it is possible to identify YFI among other plants using a four-band multispectral camera. It is, therefore, important to assess the potential role of UAV-based remote sensing in traditional approaches, although satellite and manned aircraft are limited in use. In addition, the cost of manned aircraft and high-resolution satellite images, limitations imposed by cloud cover, the satellite time and spatial resolutions are often inadequate for small patches. Thus, the different types of remote sensing provide complementary information to the IAS mapping.

The main limitations of remote sensing are linked to the high cost of very high spatial resolution and multi-date satellite images, and a dearth of these images at optimal phenological periods of detection\cite{dash2019taking}. Thus, the development and proliferation of UAVs present an opportunity for IAS research. The flexibility and cost‐efficiency of these crafts offer a valuable solution where high‐spatial or high‐temporal resolution remotely sensed data are required. However, the success of its application depends on the IAS to be mapped, the habitat characteristics of the system studied, the classification of the UAV configuration, analytical methods, and the limitations of each study\cite{dash2019taking}.

Another important advantage of the method developed is the easy implementation. The procedure is semi-supervised, meaning it needs just a priori knowledge about the flower's location to be seated and calibrated. Also, few data are used, and the model does not require training stages. Thus, the method present relevant advantages regarding the IAS monitoring strategy compared to artificial intelligence approaches.

As a disadvantage, the procedure is applicable when the target IAS exhibits a notable phenological characteristic, such as flowering. Other IAS with no notable phenological characteristics could not be detected by the process developed.

Studies revealing climatic change may benefit various species of IAS by increasing their geographic distributions and dynamics\cite{masters2010climate,bellard2014vulnerability}. Therefore, it 
is important to increase monitoring activities to reduce the uncertainty expected. Under this challenging scenario, UAV-based remote sensing may become a popular complementary tool for monitoring IAS.

\subsection*{Tool potential application in IAS management}
The IAS management principles are: 1) prevention and early eradication, 2) containment and control, and 3) general clearing and guiding principles \cite{IAS}.

A prevention strategy should include regular surveys and monitoring for IAS, rehabilitating disturbed areas, and preventing unnecessary disturbances. Monitoring plans must quickly identify invasive species as they are established. Regular updates on immediate threats are crucial \cite{delivering2008handbook}. 

When new invasive species are found, immediate responses should involve identifying the area for future monitoring and removing the plants manually or with suitable herbicides. Monitoring and acting swiftly is better than letting invasive plants become established.

If alien invasive plants become established, action plans need to be developed, considering infestation size, budgets, manpower, and time. Separate plans may be needed for different locations or species, using appropriate chemicals and control agents. The key is to prevent invasions from getting out of control, ensuring effective containment and minimal long-term resource use. This approach also minimizes the impact on natural systems.

\begin{figure*}[h]
 \centering
 \includegraphics[width=0.9\textwidth]{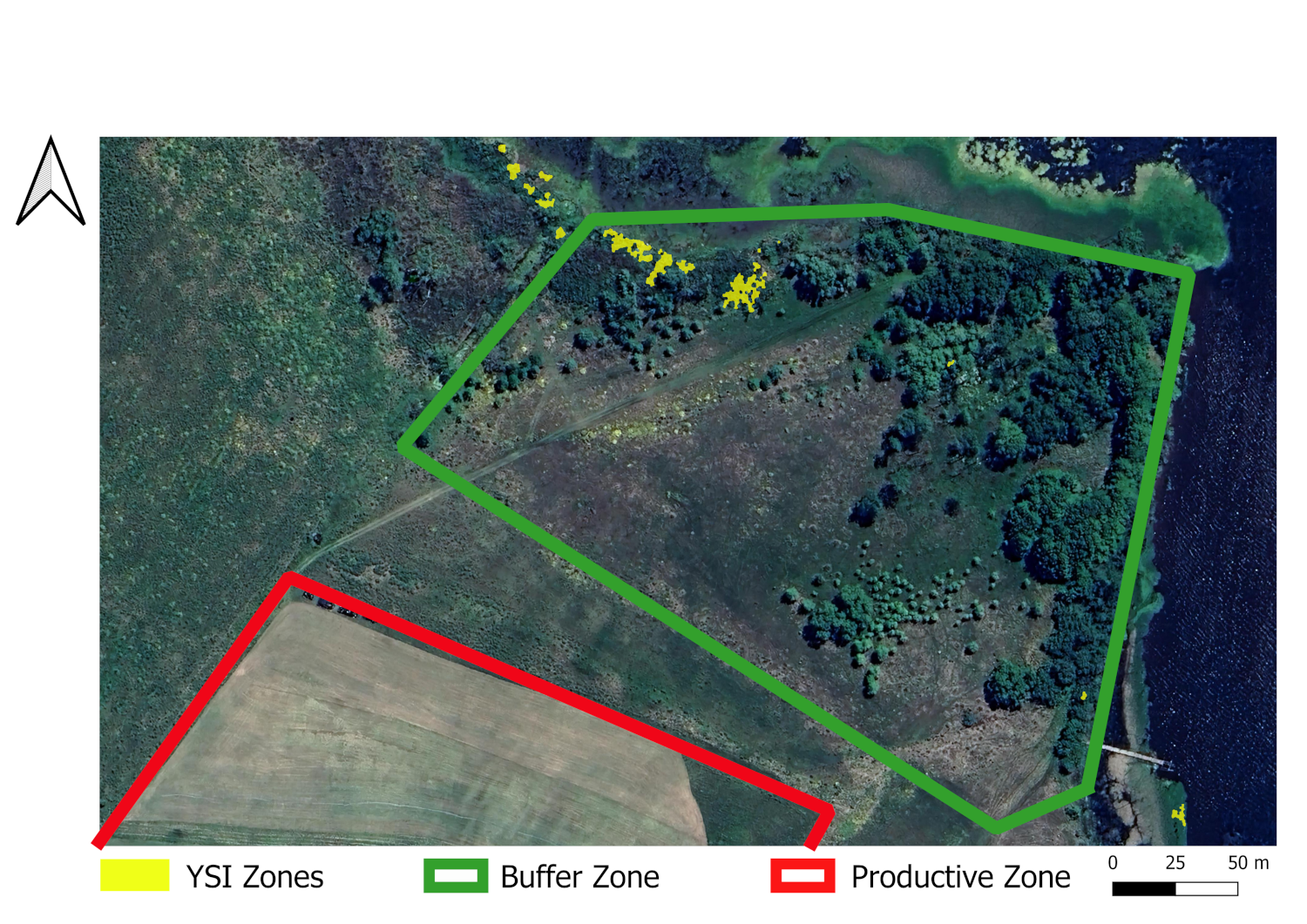}
 \caption{Example of tool application}
 \label{example}
\end{figure*}

 As shown \ref{example}, our approach can be easily integrated into the strategy, mainly in the survey and monitoring stages. Figure \ref{example} shows an example of tool application in a specific productive area. The images show the YFI, buffer zone and productive areas' locations on this field. It is helpful in detecting YFI early, allowing immediate responses before they become established. It is important to highlight that the YFI zones are geo-localized, which means a significant advantage in the IAS clearing stages. Also, this kind of images can be obtained frequently in order to analyze the YFI proliferation an the percentage of buffer zone affected by IAS.

\section*{Conclusions}

The presented approach featuring flexible UAV aerial data acquisition is at the forefront of future IAS monitoring and eradication efforts. Once the proposed semisupervised methodology is tested to produce reliable results and the UAV is fully optimized, the methods shall bring a decisive edge to IAS management policymakers, catchment management authorities, and land owners. This technology has potential for the IAS management community and can significantly contribute to the ever-growing resilience and climate change strategies.

\section*{Acknowledgments}
The research was founded by National Agency of Innovation and Investigation of Uruguay. Project: IA 2021 1 1010782. Authors thank the following individuals and organization for their assistance throughout all fieldwork: Vivero el Chajá and Mr. Franco Bascialla.

\end{document}